# BAS: An Answer Selection Method Using BERT Language Model

**Jamshid Mozafari[1], Afsaneh Fatemi[*,1], Mohammad Ali Nematbakhsh[1]**

[1] *Faculty of Computer Engineering, University of Isfahan, Isfahan, Iran*

emails: mozafari.jamshid@eng.ui.ac.ir, a_fatemi@eng.ui.ac.ir, nematbakhsh@eng.ui.ac.ir

## Abstract

In recent years, Question Answering systems have become more popular and widely used by users. Despite the increasing popularity of these systems, their performance is not even sufficient for textual data and requires further research. These systems consist of several parts that one of them is the Answer Selection component. This component detects the most relevant answer from a list of candidate answers. The methods presented in previous researches have attempted to provide an independent model to undertake the answer-selection task. An independent model cannot comprehend the syntactic and semantic features of questions and answers with a small training dataset. To fill this gap, language models can be employed in implementing the answer selection part. This action enables the model to have a better understanding of the language in order to understand questions and answers better than previous works. In this research, we will present the 'BAS' stands for BERT Answer Selection that uses the BERT language model to comprehend language. The empirical results of applying the model on the TrecQA Raw, TrecQA Clean, and WikiQA datasets demonstrate that using a robust language model such as BERT can enhance the performance. Using a more robust classifier also enhances the effect of the language model on the answer selection component. The results demonstrate that language comprehension is an essential requirement in natural language processing tasks such as answer-selection.

**Keywords:** Question answering systems, Deep learning, Answer selection, Language modeling

## 1. Introduction

Humans have always sought to find answers to their questions. Based on the type of questions they encounter, they are looking for answers (Kolomiyets & Moens, 2011). For example, for the question 'Which image is the most beautiful landscape?', the answer is an image, or for the question 'Which is the sound of the sparrow?', the answer is audio. However, it can be argued that the most common type of answers is textual. In the past, the questioner found many answers within the books. This method had two significant problems. First, all books were not readily available, and second, it took a long time to read the book and find the answer. With the advent of the Internet, resource inaccessibility problem was primarily resolved (Brill, Dumais, & Banko, 2002), but the second problem still remained. To overcome this problem, information retrieval systems and search engines have been developed. These systems receive a query from the user and return the documents containing the answer (Manning, Raghavan, & Schütze, 2008). The user could find the answer by going through these documents. The emergence of search engines was not a precise solution to the second problem because these systems returned the documents and the questioner needed to go through each of the documents in order to find the answer. To overcome the second problem, question answering systems were developed. These systems, instead of the whole document, return a word, phrase, or sentence as an exact answer.

Question answering systems are of two general types, containing Knowledge-based systems and Information retrieval-based (IR-based) systems. Knowledge-based question answering systems can be considered as a huge graph in which entities are linked through edges. Edges also represent the meaning of the relationship between entities. This information is stored in a structured manner that is extracted from the graph using query languages such as SPARQL (Perez, Arenas, & Gutierrez, 2009). The benefits of these systems include the exact answer which the system returns because it does not require text analysis, and the answer is produced through the information stored in the graph. One of the drawbacks of these systems is related to the production of knowledge graphs, as producing and implementing this huge graph is by no means an easy task and can be very time-consuming. IR-based question answering systems do not require this huge knowledge graph and attempt to extract the answer from raw texts. These systems attempt to provide a textual answer to the asked question using reading comprehension. The advantages and disadvantages of these systems are different than knowledge-based systems. Nowadays, researchers have been focusing on these systems because such systems do not need to create knowledge graphs; instead, they use raw text. IR-based question answering systems consist of four different parts, including Question Analysis, Document Retrieval, Answer Selection, and Answer Extraction (Jurafsky & Martin, 2014). Figure 1 shows the general pipeline architecture of the Information retrieval-based systems.

---

[*] **Corresponding Author**

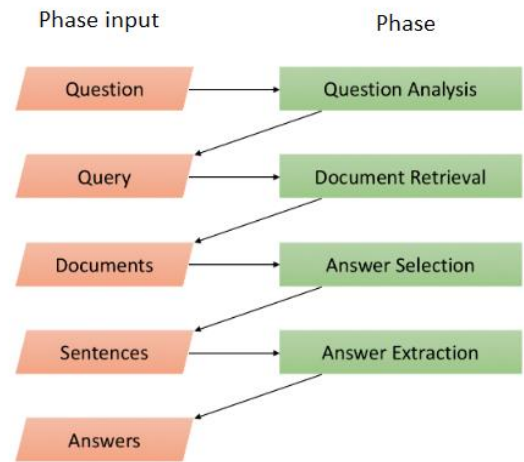

*Fig 1. The general pipeline architecture of Information retrieval-based question answering systems (Sequiera et al., 2017)*

The question analysis part receives the user question and tries to detect the answer type and generates a query for the document retrieval section. This query is passed to the document retrieval part, and documents that are more relevant to the query are retrieved, and then some of the most relevant passages are selected as retrieved passages. The answer selection part selects the most relevant from candidate sentences as the most relevant answer to the question. The answer selection part is for two general types: the extracted Answer and the generated Answer. In the extracted answer type, the answer is extracted from the passage and is passed to the answer extraction part without any changes. In the generated answer type, the answer sentence does not exist in the passage and is produced based on linguistic rules. Finally, the answer extraction part extracts the exact answer from the relevant answer sentence and returns it as the final exact answer. However, in some researches, the answer extraction part is combined with the answer selection part, and even some researches omit the answer extraction part and return the final answer as a sentence (Jurafsky & Martin, 2014; Mishra & Jain, 2016).

In the question answering domain, questions are divided into five general categories containing list type questions, hypothetical questions, causal questions, confirmation questions, and factoid questions (Mishra & Jain, 2016). Factoid questions are questions whose answer is a piece of a document, passage or sentence. In other words, the answer to the question exists in the text.

One of the most fundamental issues in natural language processing is the similarity measurement. This issue tries to identify two sentences, two paragraphs or two documents that have more similarities in terms of semantics and syntax. An essential application of similarity measurement is paraphrase recognition, which tries to identify two sentences that have the same meaning but are syntactically different (Magnolini, 2014). One of the factoid questions' features is the semantic similarity of an answer with its question. This feature makes it possible to use similarity measurement for finding answers of factoid questions in question answering systems which use the extracted answer type in answer selection part. The similarity measurement issue is a supervised classification problem. In other words, a trained classifier can predict the similarity of two sentences. Consequently, such a method can be used to respond to factoid questions. The answer-selection task can be expressed as: if $q = \{q_1, q_2, \ldots, q_n\}$ is a set of questions, for each question $q_i$, there is a candidate set of answers $\{(s_{i1}, y_{i1}), (s_{i2}, y_{i2}), \ldots, (s_{im}, y_{im})\}$ where $s_{ij}$ refers to the $j_{th}$ candidate answer for $q_i$. $y_{ij}$ also refers to the correctness of the answer, as if $y_{ij} = 1$, the answer is correct and if $y_{ij} = 0$, the answer is incorrect. If a training dataset exists that includes such information, we can train a classifier that can find the most relevant answer to factoid questions using semantic and syntactic similarities (Echihabi & Marcu, 2003; Yih, Chang, Meek, & Pastusiak, 2013). A question with three candidate answers is shown in Table 1.

*Table 1: A factoid question with three candidate answers.*

| Q | Who is the telephone inventor? | |
|---|---|---|
| $a_1$ | The first telephone was invented by Alexander Graham Bell. | $y_{11} = 1$ |
| $a_2$ | In 1875, Alexander Graham Bell succeeded in presenting the first telephone to human society. | $y_{12} = 1$ |
| $a_3$ | The first telephone was invented in 1875. | $y_{13} = 0$ |

Until now, various methods have been proposed to undertake both answer-selection and the similarity measurement tasks. These methods can be divided into two general categories. The first category is rule-based that attempted to measure the similarity between two sentences based on linguistic rules. The second category is the methods which use machine learning algorithms. In these methods, models try to learn linguistic rules automatically. In the second category, feature engineering was initially used, but in recent years, deep learning methods have become more popular, and most researchers have used it instead of feature engineering.

The deep learning-based models initially attempted to overcome the problem independently and did not use other NLP tasks such as natural language inference (Khot, Sabharwal, & Clark, 2018), paraphrase identification (Liu, He, Chen, & Gao, 2019), language modeling (Devlin, Chang, Lee, & Toutanova, 2019) and so on. However, it was revealed that answer selection methods which were combined with the other NLP tasks could provide more accurate models. In 2019, one of the topics investigated is the combination of language models with answer selection methods. In methods that have used language models, less attention has been paid to the influence of the language model and attempts to analyze the answer selection component attached to the language model (Yoon, Dernoncourt, Kim, Bui, & Jung, 2019). In this paper, we show that if a more robust language model is used, it is not required to use other tasks of NLP in answer selection task. We propose a model using a preprocessing section and various neural networks stacked on the BERT model (Devlin et al., 2019) that finds the most relevant answer to a factoid question from candidate answers. The empirical results demonstrate the superiority of our proposed model, which achieve state-of-the-art performance for TrecQA raw (M. Wang, Smith, & Mitamura, 2007), TrecQA clean (Z. Wang & Ittycheriah, 2015) and WikiQA (Y. Yang, Yih, & Meek, 2015) datasets.

In this research, we endeavour to solve the following three research questions. Our experiments also aim to solve these questions:

- Can the BAS model outperform the baseline models?
- Does the preprocessing have a significant effect on the performance of the BAS model?
- How do different classifiers affect the performance of the BAS model?

The contribution of this research paper includes:

- We propose the BAS (BERT Answer Selection model) that ranks the candidate answers in terms of semantic and syntactic similarity, using language models.
- The preprocessing increases the importance of EAT-type (Expected Answer Type) entities to find correct answers.
- The BERT language model is used to capture the meaning of input sentences better than ordinary neural networks.
- The MAP and MRR measures of the BAS model show that it performs better than state of the art.

In section 2, related works will be explained. In section 3, the proposed model will be described in detail. In section 4, the baseline models, the datasets, and implementation details will be presented. In section 5, the proposed model will be evaluated, and the results of the experiments will be discussed. Finally, the paper will be concluded in section 6.

## 2. Related Works

This section consists of two parts: in the first, we discuss the answer selection related works, and in second, the BERT language model will be briefly examined.

## 2.1 Answer Selection

Research history in the answer selection field can be divided into three different parts: the first includes the works that used lexical features, the second includes those that used feature engineering techniques, and the most recent part includes researches that used deep learning and deep neural networks.

### 2.1.1 Feature Engineering

The researches presented in the first period used the question and answer overlap; that is, the most relevant answer was selected based on the common words between the two sentences. During this period, researches used bag-of-words and bag-of-grams methods (Wan, Dras, Dale, & Paris, 2006). Some methods also used the weighted bag-of-words. For example, the question and candidate answers presented in Table 1, indicate that using these methods is not sensible (Surdeanu, Ciaramita, & Zaragoza, 2008). The weakness of these methods was to not use semantic and linguistic features of sentences (Mozafari, Nematbakhsh, & Fatemi, 2019). That's why some studies used lexical resources such as WordNet (Miller, 1998) to overcome the semantic problem, but these researches failed to remove language constraints because some words were not mentioned in these lexical resources (Tu, 2018).

The researches presented in the second period attempted to use feature engineering. Some researches used syntactic and semantic structures of sentences. For example, Punyakanok (Punyakanok, Roth, & Yih, 2004) used the dependency tree of the sentences. Other researches developed more robust models for answer-selection task using dependency tree methods and tree edit distance algorithms (Heilman & Smith, 2010; M. Wang & Manning, 2010; Yao, Durme, Callison-Burch, & Clark, 2013). Yih et al. (Yih et al., 2013) show the use of external tools such as WordNet and Named entity recognition (NER) (Jurafsky & Martin, 2014) with dependency trees, caused that semantic features were more employed. Finally, Severyn et al. (Severyn & Moschitti, 2013) presented a framework that performed feature engineering automatically and attempted to eliminate feature engineering problems to some extent. This framework can be considered one of the first attempts to eliminate feature engineering.

Nevertheless, the third period can be called the best period for answer-selection task and question answering systems because the speed of enhancing the performance of the models presented in this period is far fast than the preceding ones. This period, also called the artificial intelligence explosion, owes to the emergence of deep neural networks and deep learning. The models presented in this period utilize deep neural networks, which eliminates the need for feature engineering. These models need substantial training data. This need is a significant challenge and is resolved hardly. Due to the vast number of researches in this period, the researches are divided into five different categories, including Siamese-based, Attention-based, Compare-Aggregate-based, Language model-based, and specific methods. Each of these categories will be explained below.

### 2.1.2 Siamese-based models

The proposed Siamese-based models are models that follow the structure of the Siamese network (Bromley et al., 1993) and process questions and answers independently and provide a vector representation for each sentence. In these models, the information of the other sentence is not employed during the processing of each sentence (Lai, Bui, & Li, 2018).

Yu et al. (Yu, Hermann, Blunsom, & Pulman, 2014) presented the first model which used the deep neural network to overcome the answer-selection task. This model selects the most relevant answer from candidate answers using a convolutional neural network and logistic regression. Feng et al. (Feng, Xiang, Glass, Wang, & Zhou, 2015) used the model presented by Yu et al. They attempted to produce various models that were produced by combining deep neural networks and fully-connected networks. In these models, various types of hidden layers, convolution operations, pooling and activation functions were used. However, these models were independent and were evaluated separately. In this regard, He et al. (He, Gimpel, & Lin, 2015) developed a model that combined various models and produced a single model. They tried to produce a vector representation for each sentence. These vectors resulted from the processing of various models. The ranking method of previous models was a pointwise ranking, but Rao et al. (Rao, He, & Lin, 2016) showed that in case of using the pairwise ranking, the performance of the model is enhanced instead. In this research, a model was presented which converted each pointwise model into a pairwise model. In this regard, the model presented by He et al. (He et al., 2015) was given as a pointwise model to the Rao et al. model, which enhanced the performance of the model. Madabushi et al. (Tayyar Madabushi, Lee, & Barnden, 2018) provided a pre-processing operation rather than enhancing previous models. In their research, the named entities in candidate answers that are equivalent to the answer type announced by the question processing part, are replaced with a special token, which makes it easier for models to find the most relevant answer. This preprocessing was applied to the model presented by Rao et al. (Rao et al., 2016) and confirmed its effectiveness. The problem was to replace all the tokens with a unique one. In this regard, Kamath et al. (Kamath, Grau, & Ma, 2019), instead of replacing all named entities with a unique token, replaced each named entity with a special token. However, unlike Madabushi et al. (Tayyar Madabushi et al., 2018), they did not apply the idea to one of the previous models; instead, they presented a new model with recurrent neural networks.

### 2.1.3 Attention-based models

The proposed attention-based models are those that, unlike Siamese-based models, use context-sensitive interactions between sentences. In these models, the attention mechanism (Bahdanau, Cho, & Bengio, 2015) is used. The attention mechanism was first used in machine translation researches but later in other fields of natural language processing such as question answering and answer-selection task (Lai et al., 2018) was also used.

Yang et al. (L. Yang, Ai, Guo, & Croft, 2016) presented one of the first models which used the attention mechanism for the answer-selection task. In this research, the attention mechanism, proposed by Bahdanau et al. (Bahdanau et al., 2015) and implemented with recurrent neural networks, was used to overcome the answer-selection task. The first attention mechanism was presented only for recurrent neural networks, but He et al. (He, Wieting, Gimpel, Rao, & Lin, 2016) could provide a model for answer-selection task which used the attention mechanism in convolutional neural networks. This research showed that the combination of the attention mechanism with convolutional neural networks is more efficient than the combination of the attention mechanism with recurrent neural networks. Mozafari et al. (Mozafari et al., 2019) showed that using feature vectors, convolutional neural networks and pairwise ranking algorithms in the model presented by He et al. (He et al., 2015) can provide a more robust model.

### 2.1.4 Compare-Aggregate-based models

The proposed compare-aggregate models are models that focus on context-interaction between sentences more than attention-based models. These models first compare smaller units of sentences such as words to capture more information. Then, they aggregate the information obtained from the comparison between words and present a vector representation for each sentence (Lai et al., 2018).

He et al. (He & Lin, 2016) presented one of the first models to overcome the answer-selection task using compare-aggregate methods. Instead of converting the input sentences into a vector representation and measuring the similarity of the two sentences using the vectors, they compared the word vectors to each other and produced the vector representations of each input sentence by aggregating these values. Wang et al. (S. Wang & Jiang, 2017) used the idea of He et al. (He & Lin, 2016) and presented a general 'compare-aggregate' framework which provided excellent performance for the answer-selection task. Wang et al. (Z. Wang, Hamza, & Florian, 2017) developed this framework and showed that if two sentences are matched in two directions, and instead of word-by-word matching, each word is matched with all the components of the other sentence, a more robust model is presented. Bian et al. (Bian, Li, Yang, Chen, & Lin, 2017) used dynamic-clip technique rather than a simple attention mechanism in the 'compare-aggregate' framework and showed that this modification eliminates ineffective information and provide a more robust vector representation. Shen et al. (G. Shen, Yang, & Deng, 2017) introduced an inter-weight layer and tried to set a weight to each word. Tran et al. (Tran et al., 2018) inspired by the Additive Recurrent Neural Network (Lee, Levy, & Zettlemoyer, 2017), introduced a new recurrent neural network which understood input text content more than previous models.

### 2.1.5 Language Model-based models

The proposed language model-based models are models that instead of overcoming the answer-selection task from scratch, use pre-trained language models that have a complete understanding of the language. These models used the pre-trained language models to overcome the answer-selection task in a similar way proposed by Howard et al. (Howard & Ruder, 2018).

Yoon et al. (Yoon et al., 2019) developed a model which used language models for answer-selection task. This model used the ELMo language model (Peters et al., 2018) along with techniques such as Latent-Clustering and demonstrated that the combination of these components produced a robust model.

### 2.1.6 Special models

Some models were not in line with earlier models and tried to provide an independent model. These researches tried to create a new path in the answer selection field. However, it did not get much attention.

Wang et al. (Z. Wang, Mi, & Ittycheriah, 2016) utilized dissimilar components of the input sentences alongside similar components. They believed that dissimilar components were crucial as much as similar components in identifying the semantic similarity of sentences. Shen et al. (Y. Shen et al., 2018) developed the KABLSTM model, which utilizes knowledge graphs. They developed a context-knowledge interactive learning architecture, which used interactive information from input sentences and knowledge graph. Yang et al. (R. Yang, Zhang, Gao, Ji, & Chen, 2019) presented the RE2 model which attempted to provide a lightweight model with satisfactory performance. The model's name stands for Residual vectors, Embedding vectors and Encoded vectors. In Table 2, the related works for various datasets are shown.

*Table 2: Related works according to their characteristics.*

| Reference | Architecture | MAP TrecQA Raw | MRR TrecQA Raw | MAP TrecQA Clean | MRR TrecQA Clean | MAP WikiQA | MRR WikiQA |
|---|---|---|---|---|---|---|---|
| (Punyakanok et al., 2004) | Feature Engineering | 0.419 | 0.494 | - | - | - | - |
| (Heilman & Smith, 2010) | Feature Engineering | 0.609 | 0.692 | - | - | - | - |
| M. Wang & Manning, ) (2010 | Feature Engineering | 0.595 | 0.695 | - | - | - | - |
| (Yao et al., 2013) | Feature Engineering | 0.631 | 0.748 | - | - | - | - |
| (Yih et al., 2013) | Feature Engineering | 0.709 | 0.770 | - | - | - | - |
| Severyn & Moschitti, ) (2013 | Feature Engineering | 0.678 | 0.736 | - | - | - | - |
| (Yu et al., 2014) | Siamese | 0.711 | 0.785 | - | - | - | - |
| (Feng et al., 2015) | Siamese | 0.711 | 0.800 | - | - | - | - |
| (He et al., 2015) | Siamese | 0.762 | 0.830 | 0.777 | 0.836 | - | - |
| (Rao et al., 2016) | Siamese | 0.780 | 0.834 | 0.801 | 0.877 | 0.709 | 0.723 |
| Tayyar Madabushi et ) (al., 2018 | Siamese | 0.836 | 0.862 | 0.864 | 0.903 | - | - |
| (Kamath et al., 2019) | Siamese | **0.850** | **0.892** | - | - | 0.689 | 0.709 |
| (L. Yang et al., 2016) | Attention | 0.750 | 0.811 | - | - | - | - |
| (Mozafari et al., 2019) | Attention | 0.806 | 0.852 | - | - | - | - |
| (He & Lin, 2016) | Compare Aggregate | 0.758 | 0.821 | - | - | 0.709 | 0.723 |
| (S. Wang & Jiang, 2017) | Compare Aggregate | - | - | - | - | 0.743 | 0.754 |
| (Z. Wang et al., 2017) | Compare Aggregate | - | - | 0.801 | 0.877 | 0.743 | 0.755 |
| (Bian et al., 2017) | Compare Aggregate | - | - | 0.821 | 0.899 | 0.754 | 0.764 |
| (G. Shen et al., 2017) | Compare Aggregate | - | - | 0.822 | 0.889 | 0.733 | 0.750 |
| (Tran et al., 2018) | Compare Aggregate | - | - | 0.829 | 0.875 | - | - |
| (Yoon et al., 2019) | Language Model | - | - | 0.868 | 0.928 | 0.764 | **0.784** |
| (Z. Wang et al., 2016) | Special | - | - | 0.771 | 0.845 | 0.705 | 0.722 |
| (Y. Shen et al., 2018) | Special | 0.792 | 0.844 | 0.803 | 0.884 | 0.732 | 0.749 |
| (R. Yang et al., 2019) | Special | - | - | - | - | 0.745 | 0.761 |

### 2.2 BERT language model

With the advent of deep learning, one of the issues which has received much attention in recent years is the development of models that attempt to comprehend languages (Peters et al., 2018). These researches present a model that learns the syntactic and semantic rules of language in a variety of methods, such as next word prediction, next sentence prediction, masked word prediction (Devlin et al., 2019). In other words, this model learns a language and can produce new texts with correct syntax and semantic rules. One of the novel language models which can overcome all other language models is the BERT model (Devlin et al., 2019). This model has taken advantage of the idea presented in Transformers (Vaswani et al., 2017), which is now widely used in the natural language processing community. The BERT model will be described in more detail below.

**2.2.1 Transformer**

One of the architectures which that for machine translations is the encoder-decoder architecture (Sutskever, Vinyals, & Le, 2014). Since then, this architecture has been used as one of the most widely used architectures in machine translations. Based on this architecture, the Transformer was introduced (Vaswani et al., 2017) in which a self-attention technique was used instead of using a recurrent neural network in the encoder and decoder. The method used by the Transformer went beyond machine translations and was employed in various natural language processing tasks. One of these tasks is language models such as BERT language model, which uses the transformer encoder component to implement the language model. Figure 2 shows the transformer encoder architecture.

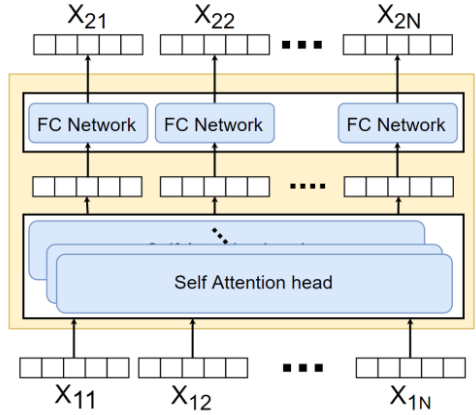

*Fig 2: Transformer encoder which consists of self-attention heads and fully connected neural networks. This encoder modifies the representation of each token to suit the contents of the other tokens and presents a new representation. Each self-attention head discovers a new semantic relation between various tokens and converts it into a new vector similar to input vectors using a fully connected neural network.*

The first step of the Transformer encoder is Self Attention. In this step, three vectors are created for each input vector X named Query, Key, and Value. The learned matrices $W_{Qi} \in \mathbb{R}^{|X| \times |Qi|}$, $W_{Ki} \in \mathbb{R}^{|X| \times |Ki|}$, and $W_{Vi} \in \mathbb{R}^{|X| \times |Vi|}$ are employed to produce Query, Key, and Value vectors for $i^{th}$ self-attention respectively:

$$Q_i = X \times W_{Qi} \tag{1}$$

$$K_i = X \times W_{Ki} \tag{2}$$

$$V_i = X \times W_{Vi} \tag{3}$$

The output vector of $i^{th}$ self-attention which is generated as follows:

$$Z_i = \sigma\left(\frac{Q_i \times K_i^T}{\sqrt{|K_i|}}\right) \times V_i \tag{4}$$

According to Figure 2, there are $|S|$ self-attention in the Transformer encoder. By concatenating the outputs of self attentions, the $Z_{1..|S|}$ vector is generated. This vector is multiplied by the learned matrix $W_O \in \mathbb{R}^{|Z_{1..|S|}| \times |H|}$ and Z vector is produced:

$$Z = Z_{1..|S|} \times W_O \tag{5}$$

Finally, the Z vector is transferred to a fully connected layer and a new vector $X_{new}$ is produced. $W_F \in \mathbb{R}^{|H| \times |X|}$ is a matrix that is equivalent to the hidden layer parameters, and $b_F \in \mathbb{R}^{|X|}$ is a vector that is equivalent to the bias:

$$X_{new} = Z \times W_F + b_F \tag{6}$$

**2.2.2 BERT Model**

The BERT language model consists of several transformer encoders stacked together. Two general types are defined based on the number of stacked encoders (L), the hidden layer size (H), and the number of self-attention heads (A). These two general types include the BERT-base and the BERT-large. The characterizes of these models are shown below:

- BERT-base: L:12, H: 768, A: 12, Training parameters: 110 M
- BERT-large: L:24, H: 1024, A: 16, Training parameters: 340 M

### 2.2.3 Fine-Tuning

Fine-tuning is to train models already trained for a particular task in order to be used for another task. The fine-tuning is used when the captured knowledge by another model needs to be used for another task. In addition to the language model presented in BERT (Devlin et al., 2019), fine-tuning has also been performed for different tasks. These tasks include Sentence Pair Classification Tasks, Single Sentence Classification Tasks, Question Answering Tasks, and Single Sentence Tagging Tasks. This paper demonstrates that the BERT language model has a high comprehension of language because, in most other tasks, it performs better than other models (Devlin et al., 2019).

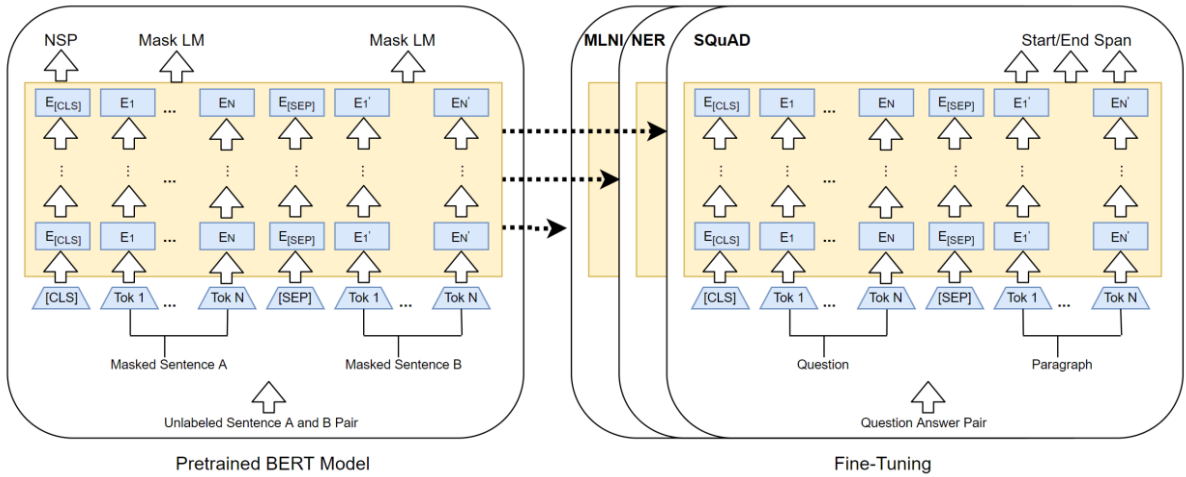

*Fig 3: BERT language model fine-tuned for various other tasks (Devlin et al., 2019).*

## 3. Proposed Model Architecture

In this paper, we present the BERT Answer Selection (BAS) Model which consists of three various sections, including preprocessing, language model, and classifier. The first section processes the input sentences and passes the processed sentences to the next section. The second section passes the processed sentences to the pre-trained language model and passes the vectors which capture the meaning of the sentences to the next section. The third section uses these vectors and performs classification. Figure 4 shows the BAS model architecture.

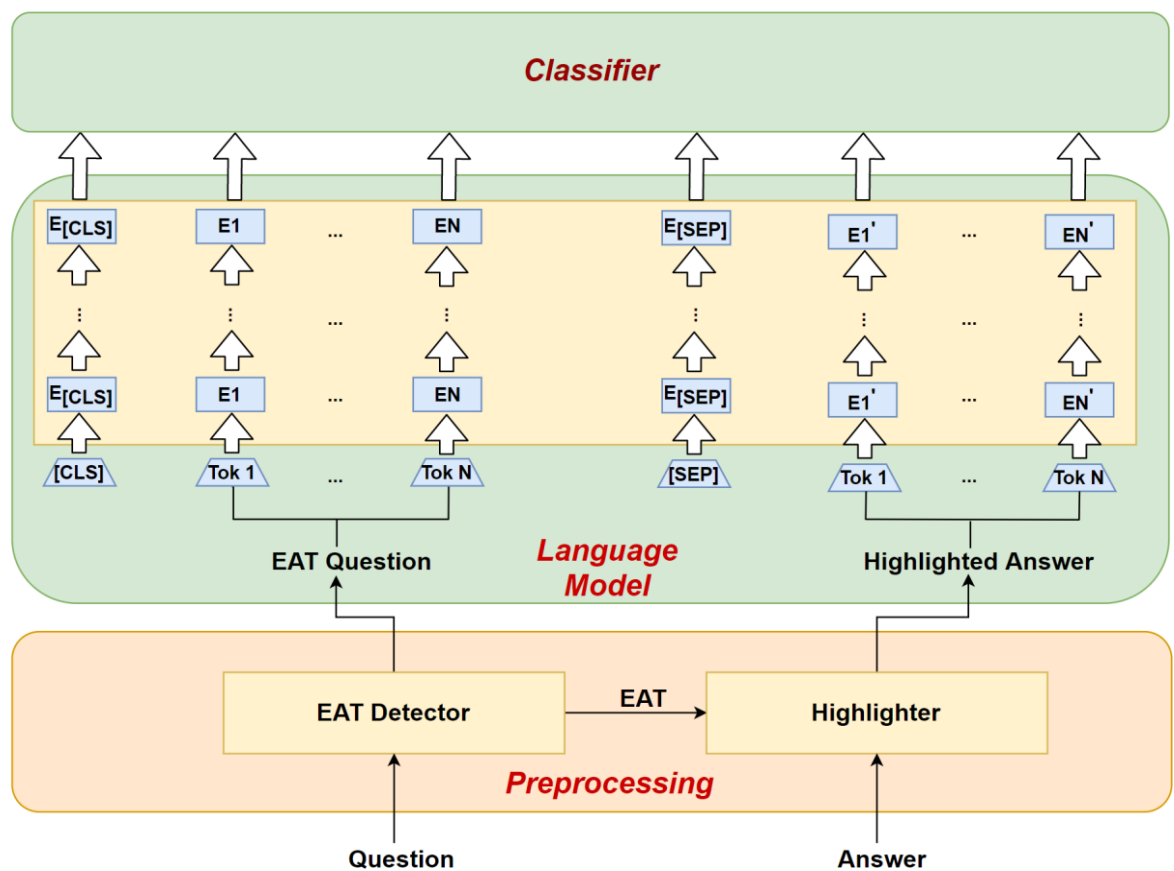

*Fig 4: BERT Answer Selection Model Architecture. The green sections are trainable and the orange section is non-trainable.*

As mentioned above, the BAS model consists of three sections: preprocessing, language model, and classifier. The preprocessing receives a question sentence and an answer sentence as input. In this section, the question is first given to the Expected Answer Type (EAT) Detector. This component detects the answer type of question and passes it to the Highlighter and annotates the question. The Highlighter component replaces all the named entities whose type is the EAT, with a special token. The processed question and processed answer sentences are then passed to the language model section. This section uses the BERT language model (Devlin et al., 2019). The question and answer are tokenized and transformed into the appropriate template for the BERT model. In this section, instead of using the BERT language model, the Sentence Pair Classification model is used. This model is known as BertForSequenceClassification. The BERT model processes the inputs and, for each token, outputs a new vector representation that captures more information from other tokens. These vectors are passed as input to the classifier section. In this section, various types of classification are employed containing classification with the fully connected neural network, classification with bag-of-words, classification with a convolutional neural network, and classification with a recurrent neural network. We will explain each section in detail below.

## 3.1 Preprocessing

In factoid questions, the exact answer to a question is a word which is appeared in the answer sentence. For example, the answer to the question 'Who is the telephone inventor?' is a sentence referring to 'Alexander Graham Bell'. For example, the sentences 'The first telephone was invented by Alexander Graham Bell' and 'In 1875, Alexander Graham Bell succeeded in presenting the first telephone to human society' are both correct answers to this question. However, the exact answer is a human name. In other words, the exact answer to the question is 'Alexander Graham Bell'. For better understanding, for example, the answer 'The first telephone was invented in 1875' is not a correct answer to the question because Alexander Graham Bell is not mentioned. As a result, a correct answer must contain a human name. More generally, the correctness probability of a candidate answer which contains named entities whose type is EAT is more than other candidates (Tayyar Madabushi et al., 2018). Earlier question answering systems process questions and answers without any preprocessing. In these systems, there is no guarantee that the system can automatically detect the answer type and selects sentences containing EAT. Madabushi et al. (Tayyar Madabushi et al., 2018) proposed a solution to this problem. They told that each candidate answer was processed separately, and if the candidate answer included EAT, replaced it with a special token. This action causes the system learns that assigns more likelihood to the sentences which contain the special token. As a result, the system can rank the candidate answers better. This idea is also used in the BAS model. To perform this, two components are needed: 'Expected Answer Type Detector' and 'Highlighter'. Each of these components will be explained below.

### 3.1.1 Expected Answer Type Detector

This component detects the answer type of questions. To perform this, we use the application programming interface (API) provided by Madabushi et al. (Tayyar Madabushi et al., 2018). Only the coarse-level of the API output is used. For example, for the question 'Who is the telephone inventor?', the answer type of the question is (HUM, ind) that coarse-level answer (HUM) is kept and the fine-level answer (ind) is discarded.

### 3.1.2 Highlighter

This section replaces EAT words of candidate answers with a special token. To perform this, named entities type of candidate answers is detected using the Spacy NER tool. Then, the detected named entities are replaced with a special token, if their type is equal to EAT. The mapping between the named entity type detected by the Spacy NER tool and the output of the EAT detector is presented in Table 3.

*Table 3: Mapping between the named entity type and the output of the expected answer type component (Kamath et al., 2019).*

| EAT | Spacy annotated tag |
|---|---|
| **HUM** | PERSON, ORG, NORP |
| **LOC** | LOC, GPE |
| **ENTY** | PRODUCT, EVENT, LANGUAGE, WORK OF ART, LAW, FAC |
| **NUM** | DATE, TIME, PERCENT, MONEY, QUANTITY, ORDINAL, CARDINAL |

The following steps describe preprocessing steps 'Who is the telephone inventor?' and 'The first telephone was invented by Alexander Graham Bell.'.

I) $Who\ is\ telephone\ inventor? \xrightarrow{EAT\ Detector} (Who\ is\ telephone\ inventor?, HUM)$

II) $The\ first\ telephone\ was\ invented\ by\ Alexander\ Graham\ Bell. \xrightarrow{Spacy\ NER\ tool}$

$The\ first\ [PRODUCT]\ was\ invented\ by\ [PERSON] \xrightarrow{Replacing\ Spacy\ annotated\ tag\ with\ EAT}$

$The\ first\ ENTY\ was\ invented\ by\ HUM \xrightarrow{Replacing\ EAT\ detector\ output\ with\ special\ token\ and\ discarding\ other\ EAT}$

$The\ first\ telephone\ was\ invented\ by\ SPECIAL\_TOKEN$

### 3.2 Language model

Texts processed by the preprocessing section are passed to the language model section. In this section, questions and answers should be transformed into an appropriate template for the BERT model (Devlin et al., 2019). In this research, we use the BERT-base language model, which has been fine-tuned for classification problems. It has a better understanding of the classification problem. The input of this model should be as follows:

$$BERT\_Input(Sentence_1, Sentence_2) = [CLS]\ Sentence_1\ [SEP]\ Sentence_2\ [SEP] \qquad (7$$

For example, for the question 'Who is the telephone inventor?' and the candidate answer 'The first telephone was invented by Alexander Graham Bell.', the input will be the following:

$$[CLS]\ Who\ is\ telephone\ inventor?\ [SEP]\ The\ first\ telephone\ was\ invented\ by\ SPECIAL\_TOKEN\ [SEP] \qquad (8$$

The [CLS] token in the BERT model is used for classification. The output of the [CLS] can be considered as a vector representation. The outputs of the [SEP] tokens do not apply to the answer-selection task and can be ignored. A new vector is presented for each token by sending this input to the language model, which captures the meaning of the token. In other words, the BERT model replaces the semantic vector of each word which independently captures the meaning of the word, with a vector that captures the meaning of the word according to its position in the sentence. The BERT model can be illustrated as follows:

$$(E_{[CLS]}, E_1, \ldots, E_N, E_{[SEP]}, E'_1, \ldots, E'_N) = BERT([CLS], Tok\ 1, \ldots, Tok\ N, [SEP], Tok\ 1', \ldots, Tok\ N') \quad (9$$

### 3.3 Classifier

In BERT paper (Devlin et al., 2019), it is suggested to use the output of the [CLS] token for classification. In this section, in addition to the classification method presented in the BERT paper, other methods will be implemented. Each of these methods will be explained below.

#### 3.3.1 BERT-base-Baseline (BB-Baseline)

In this method, the classification method proposed by Devlin et al. (Devlin et al., 2019) is employed. That is, the output of the [CLS] token, a vector of length 768, is passed as input to a fully connected neural network with a hidden layer of length 1024. The output layer of the fully connected neural network consists of two elements that the first indicates the correctness of the answer candidate, and the latter indicates the incorrectness of the answer candidate. Figure 5 presents the pseudo-code of this method. $W_{h1} \in \mathbb{R}^{1024 \times 768}$ is a matrix that is equivalent to the hidden layer parameters, and $b_{h1} \in \mathbb{R}^{1024}$ is a vector that is equivalent to the bias for the hidden layer. $W_{h2} \in \mathbb{R}^{2 \times 1024}$ is a matrix that is equivalent to the output layer parameters, and $b_{h1} \in \mathbb{R}^2$ is a vector that is equivalent to the bias for the output layer. Relu (Nair & Hinton, 2010) and Softmax (Jurafsky & Martin, 2014) activation functions are also used. Figure 6 illustrates the architecture of this method.

$BB\_BASELINE(q, a)$

$\quad Question = \boldsymbol{EAT - Detector}(q)$

$\quad Answer = \boldsymbol{Highlighter}(a)$

$\quad Input = \boldsymbol{BERT\_Input}(Question, Answer)$

$\quad (E_{[CLS]}, E_1, \ldots, E_N, E_{[SEP]}, E'_1, \ldots, E'_N, E_{[SEP]}) = \boldsymbol{BERT}(Input.TOKENS)$

$\quad H_L = \boldsymbol{relu}(W_{h1} E_{[CLS]} + b_{h1})$

$\quad f(q, a) = \boldsymbol{softmax}(W_{h2} H_L + b_{h2})$

*Fig 5: BB-Baseline model pseudo-code*

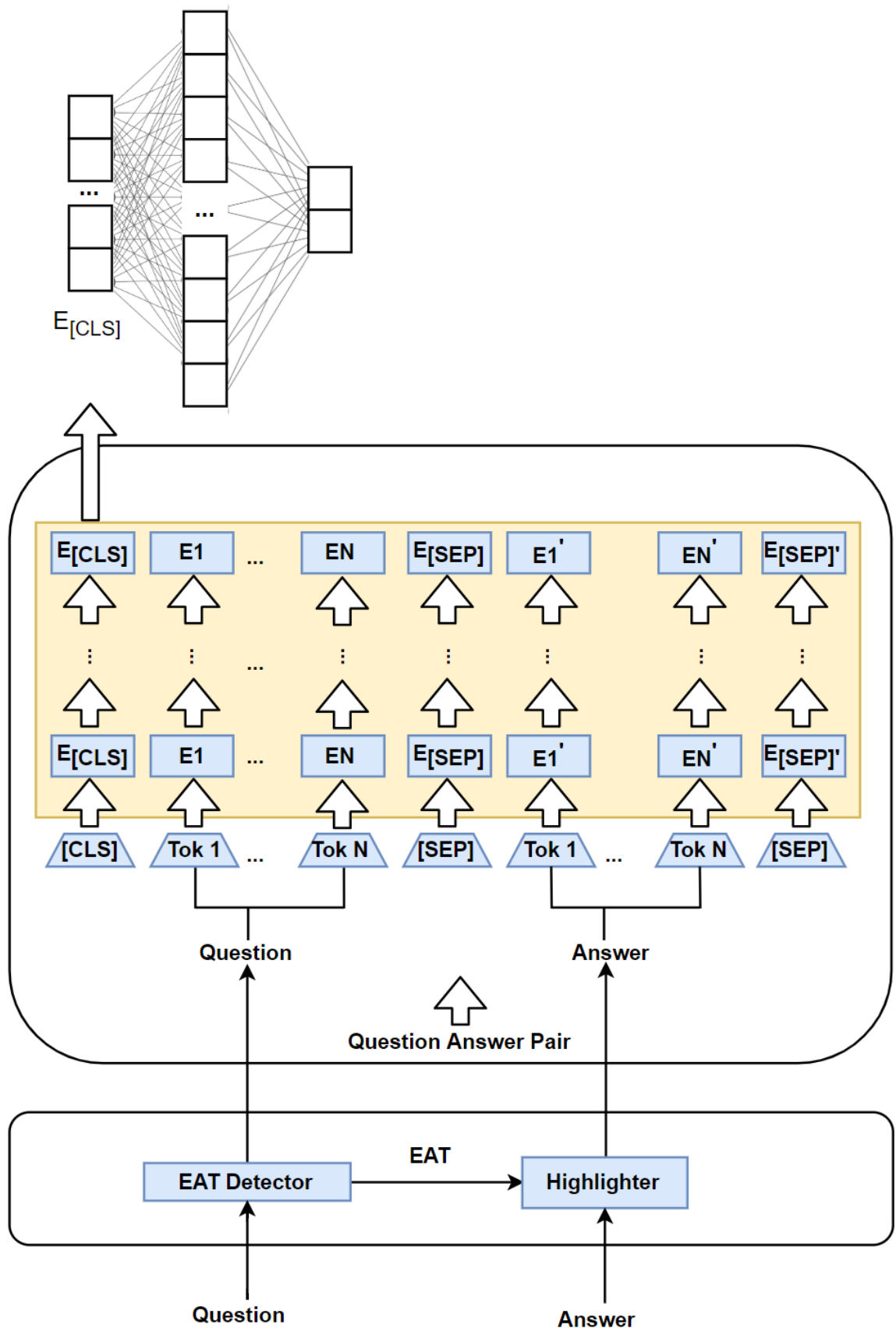

*Fig 6: BB-Baseline model architecture*

In this method, only the output of the [CLS] token is used, and the other output vectors are ignored.

### 3.3.2 BERT-base-BOW (BB-BOW)

In this method, in addition to the output vector of the [CLS] token, the output vectors of questions and answers tokens are also used for classification. That is, the token vectors of each sentence are summed, and a new vector of length 768 is presented for each sentence. As a result, there will be three vectors of length 768 which are the output vectors of question tokens, answer tokens, and the [CLS] token, respectively. A vector of length 2304 is produced by concatenating these three vectors together and is passed as input to a fully neural network connected with a hidden layer of length 1024. Figure 7 presents the pseudo-code of this method. $W_{h1} \in \mathbb{R}^{1024 \times 2304}$ is a matrix that is equivalent to the hidden layer parameters, and $b_{h1} \in \mathbb{R}^{1024}$ is a vector that is equivalent to the bias for the hidden layer. $W_{h2} \in \mathbb{R}^{2 \times 1024}$ is a matrix that is equivalent to the output layer parameters, and $b_{h2} \in \mathbb{R}^{2}$ is a vector that is equivalent to the bias for the output layer. The Concat function concatenates the input vectors and produces a matrix. Figure 8 illustrates the architecture of this method.

$BB\_BOW(q, a)$

$\quad Question = \boldsymbol{EAT - Detector}(q)$

$\quad Answer = \boldsymbol{Highlighter}(a)$

$\quad Input = \boldsymbol{BERT\_Input}(Question, Answer)$

$\quad \left(E_{[CLS]}, E_1, \ldots, E_N, E_{[SEP]}, E'_1, \ldots, E'_N, E_{[SEP]}\right) = \boldsymbol{BERT}(Input.TOKENS)$

$\quad E1\_N = \boldsymbol{concat}(E_1, \ldots, E_N)$

$\quad E1'\_N' = \boldsymbol{concat}(E'_1, \ldots, E'_N)$

$\quad BOW_1 = \sum_{i=1}^{N} E1\_N[i]$

$\quad BOW_2 = \sum_{i=1}^{N} E1'\_N'[i]$

$\quad I_L = \boldsymbol{concat}(E_{[CLS]}, BOW_1, BOW_2)$

$\quad H_L = \boldsymbol{relu}(W_{h1} I_L + b_{h1})$

$\quad f(q, a) = \boldsymbol{softmax}(W_{h2} H_L + b_{h2})$

*Fig 7: BB-BOW model pseudo-code*

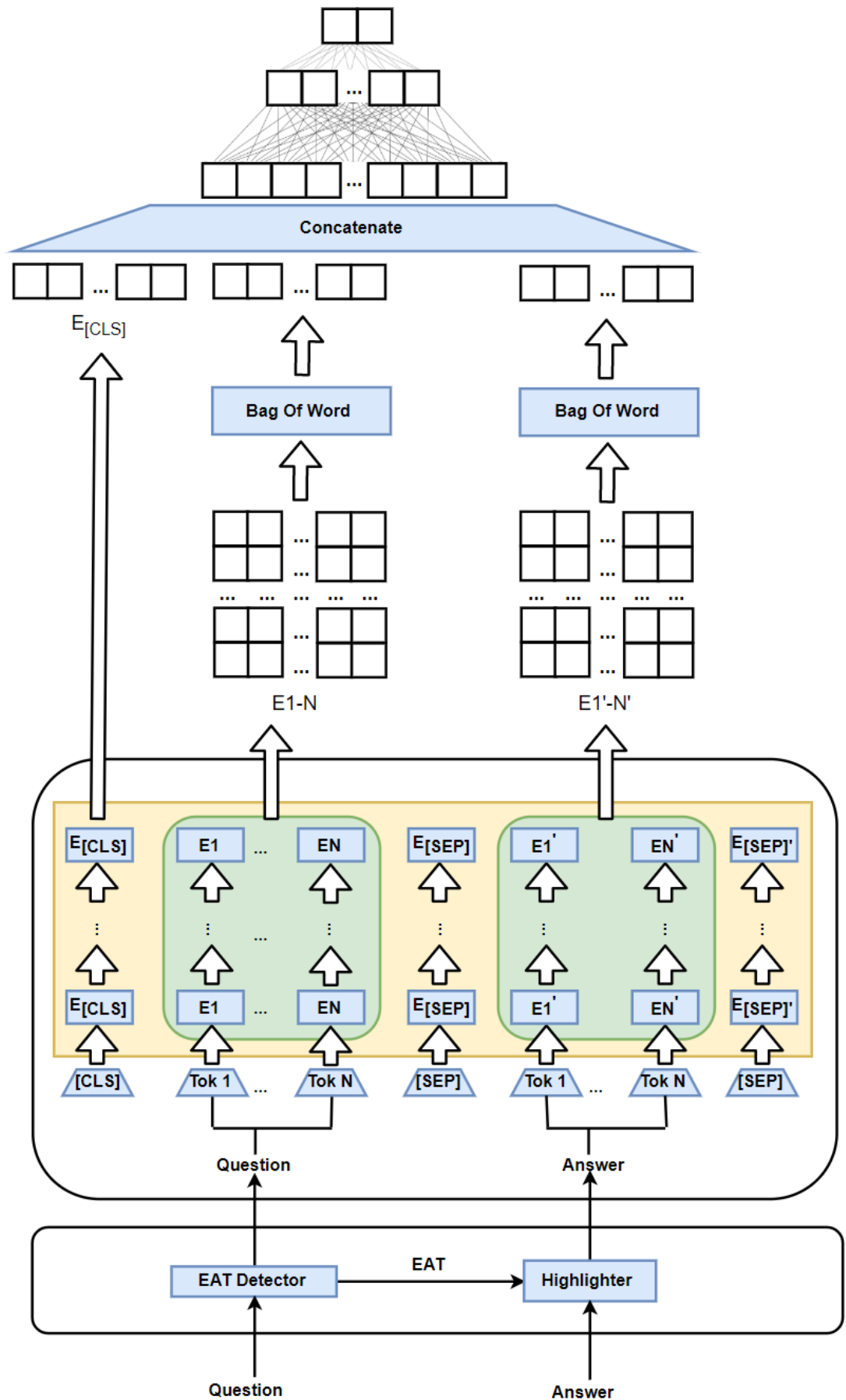

*Fig 8: BB-BOW model architecture*

### 3.3.3 BERT-base-CNN (BB-CNN)

This method also uses the output vector of the other tokens. However, the bag-of-words method is not employed; instead, the convolutional neural network is used. In this network, the window size is 3 and accordingly, the padding value is 2. The number of filters is 200, and MaxPooling is used for the pooling operation. A vector of length 200 is produced for each sentence by applying a convolutional neural network with these features. These vectors are concatenated to the output vector of the [CLS] token and a vector of 1168 lengths is produced. This vector is passed to a fully connected neural network whose hidden layer size is 1024. Then, classification operation is performed. Figure 9 is a pseudo-code of this method. $W_{h1} \in \mathbb{R}^{1024 \times 1168}$ is a matrix that is equivalent to the hidden layer parameters, and $b_{h1} \in \mathbb{R}^{1024}$ is a vector that is equivalent to the bias for the hidden layer. $W_{h2} \in \mathbb{R}^{2 \times 1024}$ is a matrix that is equivalent to the output layer parameters, and $b_{h2} \in \mathbb{R}^{2}$ is a vector that is equivalent to the bias for the output layer. The CNN function refers to the convolutional neural network. The MaxPool function also performs maximum pooling. Figure 10 illustrates the architecture of this method.

$BB\_CNN(q, a)$

$\quad Question = \boldsymbol{EAT - Detector}(q)$

$\quad Answer = \boldsymbol{Highlighter}(a)$

$\quad Input = \boldsymbol{BERT\_Input}(Question, Answer)$

$\quad (E_{[CLS]}, E_1, \ldots, E_N, E_{[SEP]}, E'_1, \ldots, E'_N, E_{[SEP]}) = \boldsymbol{BERT}(Input.TOKENS)$

$\quad E1\_N = \boldsymbol{concat}(E_1, \ldots, E_N)$

$\quad E1'\_N' = \boldsymbol{concat}(E'_1, \ldots, E'_N)$

$\quad CNN_1 = \boldsymbol{CNN}_{window_{size}=3, filters=200, padding=2}(E1\_N)$

$\quad CNN_2 = \boldsymbol{CNN}_{window_{size}=3, filters=200, padding=2}(E1'\_N')$

$\quad FeatVec_1 = \boldsymbol{MaxPool}(CNN_1)$

$\quad FeatVec_2 = \boldsymbol{MaxPool}(CNN_2)$

$\quad I_L = \boldsymbol{concat}(E_{[CLS]}, FeatVec_1, FeatVec_2)$

$\quad H_L = \boldsymbol{relu}(W_{h1} I_L + b_{h1})$

$\quad f(q, a) = \boldsymbol{softmax}(W_{h2} H_L + b_{h2})$

*Fig 9: BB-CNN model pseudo-code*

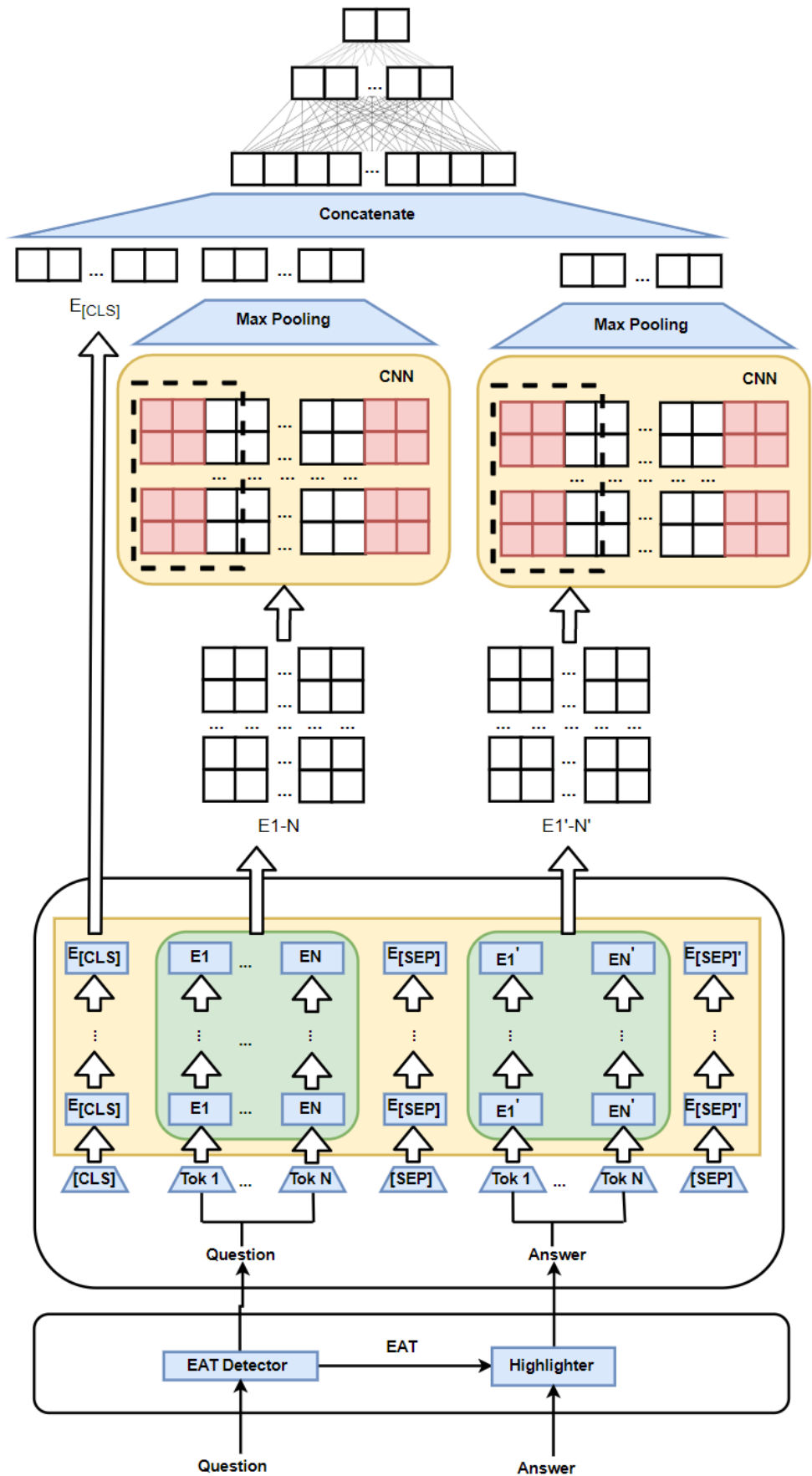

*Fig 10: BB-CNN model architecture*

### 3.3.4 BERT-base-RNN (BB-RNN)

In this method, other tokens vectors are also used. However, instead of using a convolutional neural network, it uses a recurrent neural network. The network is a two stacked RNN whose hidden layer size is 768. For each of the input sentences, a vector of length 768 is produced, and a vector of length 2304 is produced by concatenating these vectors together. This vector is passed to a fully connected neural network whose hidden layer size is 1024. Then, classification operation is performed. Figure 11 presents a pseudo-code of this method. $W_{h1} \in \mathbb{R}^{1024 \times 2304}$ is a matrix that is equivalent to the hidden layer parameters, and $b_{h1} \in \mathbb{R}^{1024}$ is a vector that is equivalent to the bias for the hidden layer. $W_{h2} \in \mathbb{R}^{2 \times 1024}$ is a matrix that is equivalent to the output layer parameters, and $b_{h2} \in \mathbb{R}^2$ is a vector that is equivalent to the bias for the output layer. The RNN function refers to the recurrent neural network. Figure 12 illustrates the architecture of this method.

$BB\_RNN(q, a)$

$\quad Question = \boldsymbol{EAT - Detector}(q)$

$\quad Answer = \boldsymbol{Highlighter}(a)$

$\quad Input = \boldsymbol{BERT\_Input}(Question, Answer)$

$\quad (E_{[CLS]}, E_1, \dots, E_N, E_{[SEP]}, E'_1, \dots, E'_N, E_{[SEP]}) = \boldsymbol{BERT}(Input.TOKENS)$

$\quad E1\_N = \boldsymbol{concat}(E_1, \dots, E_N)$

$\quad E1'\_N' = \boldsymbol{concat}(E'_1, \dots, E'_N)$

$\quad RNN_1 = \boldsymbol{RNN}_{hiddenLayer=768, N=2}(E1\_N)$

$\quad RNN_2 = \boldsymbol{RNN}_{hiddenLayer=768, N=2}(E1'\_N')$

$\quad I_L = \boldsymbol{concat}(E_{[CLS]}, RNN_1, RNN_2)$

$\quad H_L = \boldsymbol{relu}(W_{h1} I_L + b_{h1})$

$\quad f(q, a) = \boldsymbol{softmax}(W_{h2} H_L + b_{h2})$

*Fig 11: BB-RNN model pseudo-code*

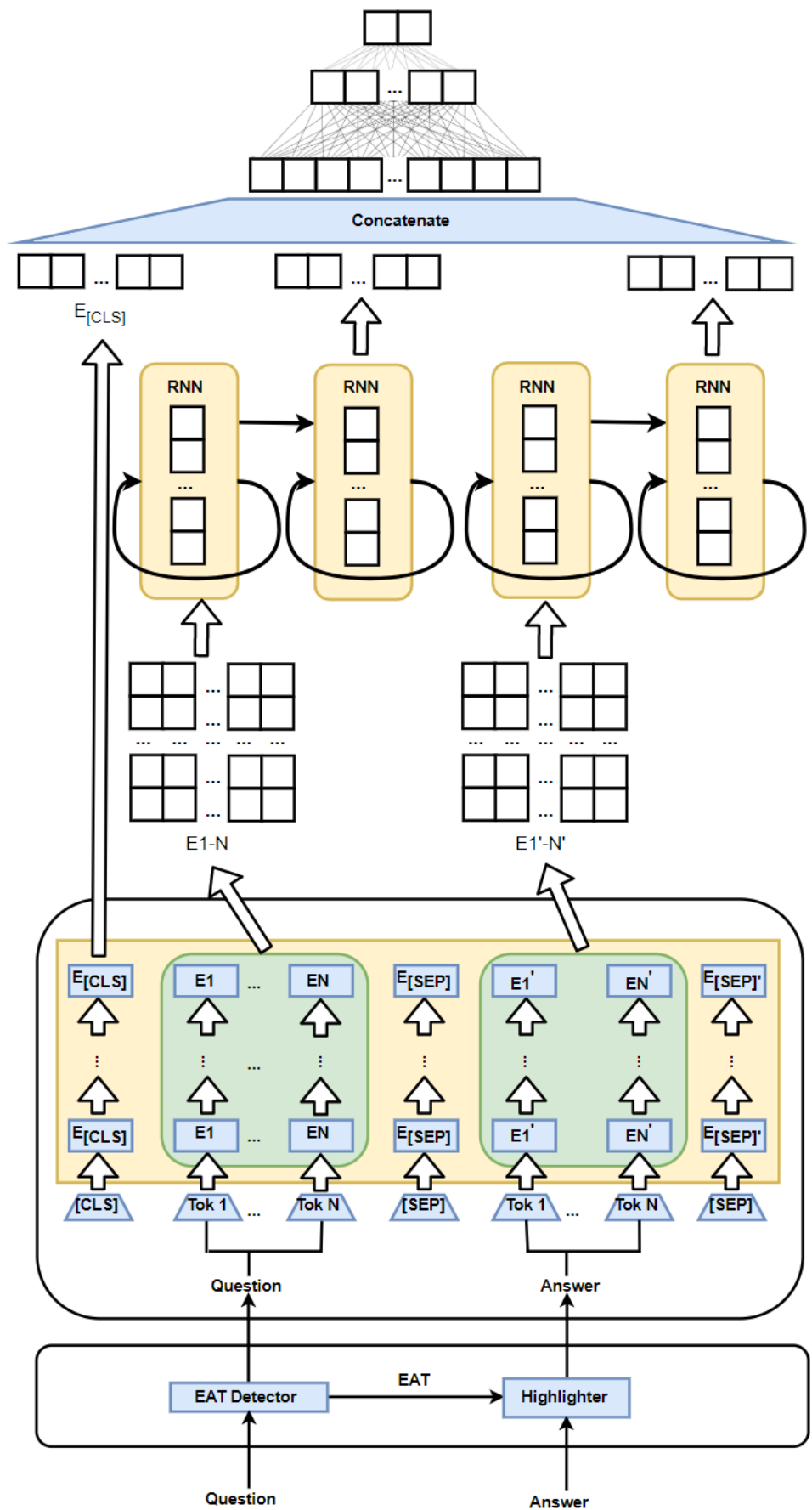

*Fig 12: BB-RNN model architecture*

### 3.4 Training Algorithm

To train the proposed model, the training parameters must be tuned so that the model enables us to find the best answer to the user's question. The training algorithm is as follows. The Q refers to the collection of questions, A refers to the collection of answers, and L refers to the labels of the answers. BAS_Model refers to one of the models, such as BB-Baseline, BB-BOW, BB-CNN or BB-RNN. Cross_entropy (Buduma & Locascio, 2017) is a loss function. Optimize function also endeavours to tune the training parameters in order to minimize loss value.

$TRAIN\_BAS(Q, A, L)$

   For each **Epoch** in {*1,2,3,4}:*

      Foreach **Batch** in $(Q, A, L)$:

         Foreach $(q, a, l)$ in $Batch$:

            $prediction = \boldsymbol{BAS\_Model}(q, a)$

            $loss_{val} = \boldsymbol{Cross\_entropy}(prediction, l)$

            $\boldsymbol{Optimize}(W_{BAS\_Model}, loss_{val})$

*Fig 13: Training algorithm*

## 4. Experiments

In this section, we briefly describe the baseline models and compare their results with the proposed model. We then explain the datasets we evaluate the proposed model with. Finally, we provide evaluation metrics and implementation details.

### 4.1 Baseline Models

To prove the superiority of the proposed model, it should be compared with some competitive baseline models. That's why, we compare the proposed model with the two competitive baseline models which have the best results. The baseline models and our proposed model are summarized in Table 4.

*Table 4: Summarization of baseline models and the proposed model*

| Architecture | Description |
|---|---|
| (Kamath et al., 2019) | A Bi-LSTM model which performs a preprocessing algorithm on the input sentences. In this preprocessing, the named entities which are equivalent to the answer type announced by the question processing part, are replaced with a special token. |
| (Yoon et al., 2019) | A model which used language models for answer-selection task. This model used the ELMo language model along with techniques such as Latent-Clustering and demonstrated that the combination of these components produced a robust model. |
| **BB-BOW** | A model which uses the output vector of the [CLS] token, the output vectors of questions and answers tokens of the BERT language model to find the best answer. That is, the token vectors of each sentence are summed, and a new vector is presented for each sentence. |
| **BB-CNN** | A model which uses the output vector of the [CLS] token, the output vectors of questions and answers tokens of the BERT language model to find the best answer. That is, the token vectors of each sentence are transferred to a CNN, and a new vector is presented for each sentence. |
| **BB-RNN** | A model which uses the output vector of the [CLS] token, the output vectors of questions and answers tokens of the BERT language model to find the best answer. That is, the token vectors of each sentence are transferred to a RNN, and a new vector is presented for each sentence. |

### 4.2 Dataset

Three datasets are used to evaluate the BAS model, including TrecQA Raw (M. Wang et al., 2007), TrecQA Clean (Z. Wang & Ittycheriah, 2015), and WikiQA (Y. Yang et al., 2015). Each of these datasets will be explained in more detail below.

### 4.2.1 TrecQA Raw

The TrecQA Raw dataset is one of the most commonly used datasets in the answer-selection task built by Yao et al. (Yao et al., 2013) from Trec Question Answering Tracks. Trec Question Answering Track 8-12 data is used to produce training data, and Trec Question Answering Track 13 data is used for validation data and test data. In this dataset, training data consist of 1229 questions and 53417 pairs, evaluation data consist of 82 questions and 1148 pairs and test data consist of 100 questions and 1517 pairs.

### 4.2.2 TrecQA Clean

The TrecQA Clean dataset is made from the TrecQA Raw dataset. In this dataset, questions that have no correct answers or only one correct/incorrect answer are removed from the validation and test data. Training data such as TrecQA Raw consists of 1229 questions and 53417 pairs. However, the validation data and test data are different from the TrecQA Raw dataset. Validation data consist of 65 questions and 1117 pairs and test data consist of 68 questions and 1142 pairs.

### 4.2.3 WikiQA

The WikiQA dataset consists of Bing search engine logs. Candidate answers to each question are extracted from Wikipedia pages. This dataset also eliminates questions that do not have the correct candidate answers. Training data consists of 873 questions and 8672 pairs, validation data consist of 126 questions and 1130 pairs, and test data consist of 243 questions and 2351 pairs.

The characteristics of these datasets are presented in Table 5.

*Table 5: Detail of the TrecQA Raw, TrecQA Clean, and WikiQA datasets*

| Dataset | Set | Number of Questions | Number of Pairs |
|---|---|---|---|
| **TrecQA RAW** | Train | 1229 | 53417 |
|  | Validation | 82 | 1148 |
|  | Test | 100 | 1517 |
| **TrecQA CLEAN** | Train | 1229 | 53417 |
|  | Validation | 65 | 1117 |
|  | Test | 68 | 1142 |
| **WikiQA** | Train | 873 | 8672 |
|  | Validation | 126 | 1130 |
|  | Test | 243 | 2351 |

### 4.3 Evaluation Metrics

MAP and MRR metrics are used in the answer-selection task to evaluate models and methods. These measures show the rating quality of candidate answers. The MRR measure only considers the rank of the first relevant answer, but the MAP measure considers the order of all relevant answers (Manning et al., 2008). These two measures are shown below.

$$MAP(Q) = \frac{1}{|Q|} \sum_{j=1}^{|Q|} \frac{1}{m_j} \sum_{k=1}^{m_j} Precision(R_{jk}) \qquad (10)$$

$$MRR(Q) = \frac{1}{|Q|} \sum_{j=1}^{|Q|} r_j \qquad (11)$$

In these equations, Q is the set of questions, $m_j$ is the number of relevant answers to $q_j$, $R_{jk}$ is a list of candidate answers that contains top k relevant answers, Precision function is a function that measures the ratio of the number of relevant answers to the total candidate answers, $r_j$ is the inverse of the first relevant answer rank for $q_j$.

### 4.4 Implementation Details

We implement the BAS model with PyTorch library (Subramanian, 2018) in Python 3.6 programming language on the Colab platform[†]. The model is trained on NVIDIA Tesla K80. We use BERT wordpiece tokenizer to tokenize input sentences. The batch size is equal to 32. We consider a vector initialized with zero vectors. The dropout is set to 0.2. Gelu (Hendrycks & Gimpel, 2016) function is used for activation function in BERT language model and Relu (Agarap, 2018) is used for fully connected layer activation function.

---
[†] https://colab.research.google.com

In BB-Baseline, BB-BOW, and BB-RNN models, the number of hidden units of the fully connected layer is equal to 1024. In BB-CNN model, the number of filters of the convolutional neural network is equal to 200. Thee window size is set to 2. The Max Pooling is applied for pooling operation.

To train the proposed model, we set the learning rate to 0.0001. The model is trained for 4 epochs. AdamW optimizer (Loshchilov & Hutter, 2019) and WarmupLinearSchedule scheduler (Devlin et al., 2019) are used for training. The BERT model used in this research is fine-tuned based on the training dataset during training.

As shown in Figure 4, the Language model and the Classifier sections are trainable and the Preprocessing section is non-trainable. The training parameters number of the Language model section is equal to 110M. In BB-Baseline, the total number of training parameters of the fully connected layer is about 768×1024+1024×2=789k. Hence the total number of training parameters in BB-Baseline is 110000k+789k≈110789k. In BB-BOW, the total number of training parameters of the fully connected layer is about 3×768×1024+1024×2=2361k. Hence the total number of training parameters in BB-BOW is 110000k+2361k≈112361k. In BB-CNN, the number of parameters in the convolution layer is about 2×768×2=3k, and the total number of training parameters of the fully connected layer is about (2×200+768)×1024+1024×2=1200k. Hence the total number of training parameters in BB-CNN is 110000k+1200k+3k≈111203k. In BB-RNN, the number of parameters in the recurrent layer is about 2×768=1k, and the total number of training parameters of the fully connected layer is about 3×768×1024+1024×2=2361k. Hence the total number of training parameters in BB-RNN is 110000k+2361k+1k≈112362k.

## 5. Results and Discussion

In this section, we explain the experiments results of the BAS model in detail. In other words, we respond to the research questions. In this regard, section 5.1 answers whether the BAS model can outperform other baseline models. Section 5.2 answers whether the preprocessing has a significant effect on the performance. Section 5.3 answers how different classifiers affect the BAS model performance.

### 5.1 Model Performance

The model presented by Kamath et al. (Kamath et al., 2019), is a model that uses the idea of appending pre-processing to answer-selection models. This model uses a preprocessing algorithm similar to the preprocessing section presented in our research. To prove their claim, they apply a simple LSTM to model the input sentences. Kamath et al. prove their claim by evaluating their model on the TrecQA Raw dataset. Currently, this model is the state-of-the-art model on the TrecQA Raw dataset.

The model presented by Yoon et al. (Yoon et al., 2019) is a model that uses the idea of using language models. It employs the ELMo language model (Peters et al., 2018). In one of their experiments, Yoon et al. investigated the effect of the lack of the language model on their answer-selection model performance and demonstrated that using the language model can improve model performance. By evaluating this model on the TrecQA Clean and the WikiQA datasets, they proved their claim. Currently, this model is the state-of-the-art model for TrecQA Clean and WikiQA datasets. The WikiQA dataset is the only dataset shared between the two baseline models. By examining the results of the baseline models on this dataset, it is proven that the idea of using language models has a more significant impact than preprocessing on the answer-selection model performance.

*Table 6: Evaluation of the proposed model*

| Architecture | TrecQA Raw | | TrecQA Clean | | WikiQA | |
|---|---|---|---|---|---|---|
| | MAP | MRR | MAP | MRR | MAP | MRR |
| (Kamath et al., 2019) | 0.850 | 0.892 | - | - | 0.689 | 0.709 |
| (Yoon et al., 2019) | - | - | 0.868 | 0.928 | 0.764 | 0.784 |
| **BB-BOW** | 0.871 | 0.898 | 0.909 | 0.946 | **0.817** | **0.835** |
| **BB-CNN** | 0.863 | 0.893 | 0.909 | 0.938 | 0.790 | 0.805 |
| **BB-RNN** | **0.872** | **0.899** | **0.915** | **0.959** | 0.784 | 0.801 |

The BAS model is compared with baselines in Table 6. The results show that the idea of using language models improves the performance of answer-selection models. The MAP and MRR metrics are increased for TrecQA Raw, TrecQA Clean, and WikiQA datasets. This proves idea of using language modeling are significantly practical. For TrecQA Raw data, the best results belong to the BB-RNN model. In this model, the MAP and MRR metrics are improved by 2.2% and 0.7%, respectively. For TrecQA Clean data, the best results also belong to BB-RNN. As shown in Table 6, the MAP and MRR metrics are improved by 4.3% and 3.1%, respectively. For WikiQA data, the best results belong to the BB-BOW model. The results show that the MAP and MRR metrics are improved by 5.3% and 5.1%, respectively. These results are shown in Figure 14 and Figure 15.

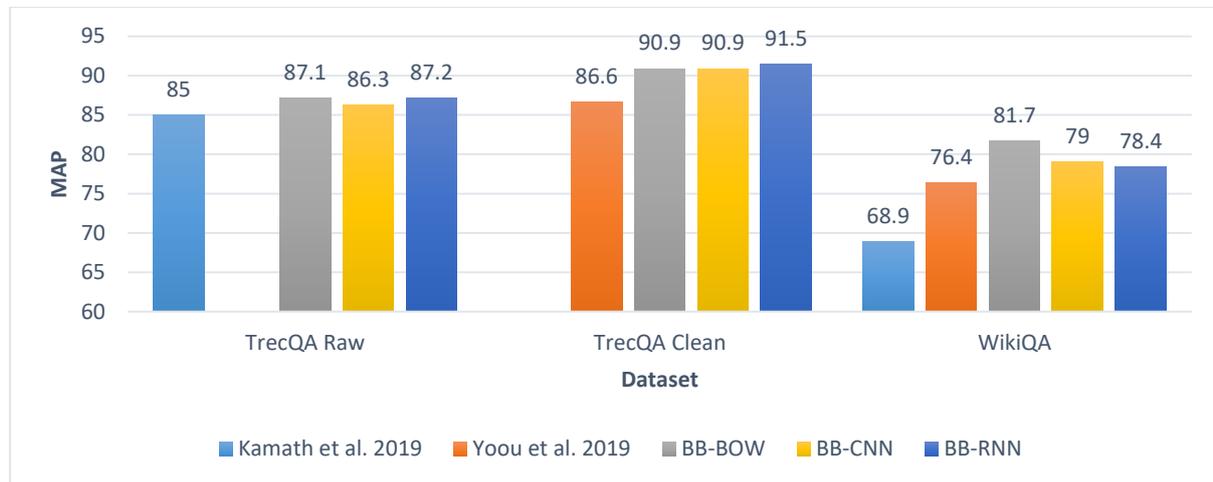

*Fig 14: Model performance on different datasets in comparison with baseline models in terms of MAP*

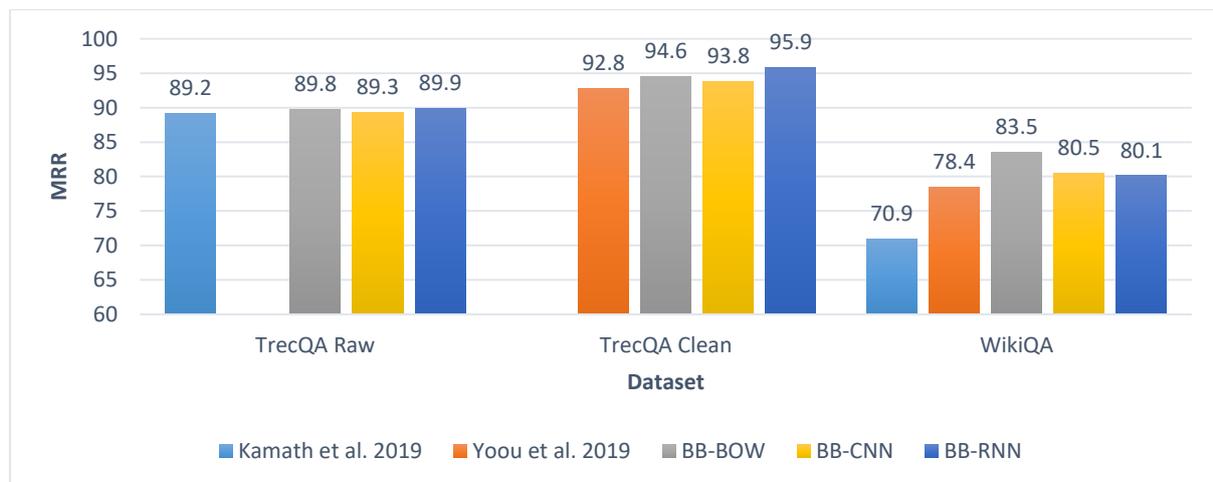

*Fig 15: Model performance on different datasets in comparison with baseline models in terms of MRR*

These results demonstrate that using a robust language model has a significant impact on the performance of the answer-selection model. This shows that just using a language model is not enough, and each language model can lead to different results. The effect of preprocessing on language model-based answer-selection model is still unclear. In the next section, we will examine the lack of the preprocessing.

## 5.2 Lack of Preprocessing

The preprocessing section attempts to replace the possible answer token contained in the candidate's answer with a special token. This allows the model to learn that the candidate answers that contain the special token are probably correct. The model presented by Kamath et al. proved this claim. But the impact of this idea on language model-based answer-selection models has not been examined so far. In this section, the preprocessing is removed from the BAS model and the results of the modified model are computed.

*Table 7: Evaluation of the effect of lack of the preprocessing (pp) on the BAS model*

| Architecture | TrecQA Raw | | TrecQA Clean | | WikiQA | |
|---|---|---|---|---|---|---|
| | MAP | MRR | MAP | MRR | MAP | MRR |
| **BB-BOW(without pp)** | 0.862 | 0.880 | 0.889 | 0.924 | 0.793 | 0.819 |
| **BB-CNN(without pp)** | 0.861 | 0.890 | 0.903 | 0.938 | 0.792 | 0.810 |
| **BB-RNN (without pp)** | 0.868 | 0.894 | 0.912 | 0.948 | 0.786 | 0.803 |
| **BB-BOW** | 0.871 | 0.898 | 0.909 | 0.946 | **0.817** | **0.835** |
| **BB-CNN** | 0.863 | 0.893 | 0.909 | 0.938 | 0.790 | 0.805 |
| **BB-RNN** | **0.872** | **0.899** | **0.915** | **0.959** | 0.784 | 0.801 |

Table 7 shows the results of the alteration. As the results demonstrate, this alteration has a negative effect on model performance. The performance of the BB-BOW model has reduced by removing the preprocessing section. However, the performance of the BB-CNN model and the BB-RNN model have fewer changes than the BB-BOW model. The reason that lack of the preprocessing has a low effect is the complexity of classifiers employed in these models. It has also reduced the BAS model performance for TrecQA Raw and TrecQA Clean datasets. In contrast, the performance of the BB-CNN and BB-RNN models for WikiQA dataset has improved slightly. The reason is the sparsity of training data in WikiQA dataset because the model has not been able to properly learn the effect of the special token on finding the correct answer. These results are shown in Figure 16 and Figure 17.

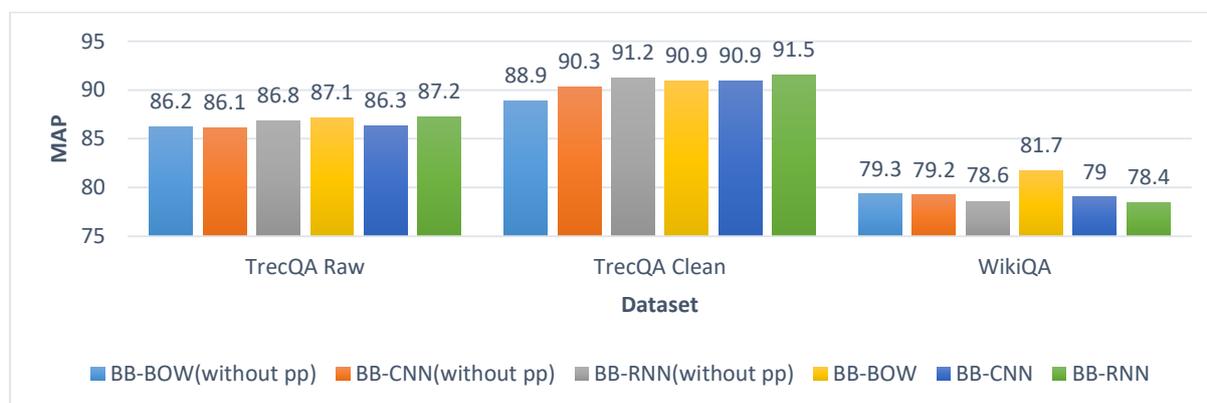

*Fig 16: Effect of lack of the preprocessing on different datasets in terms of MAP*

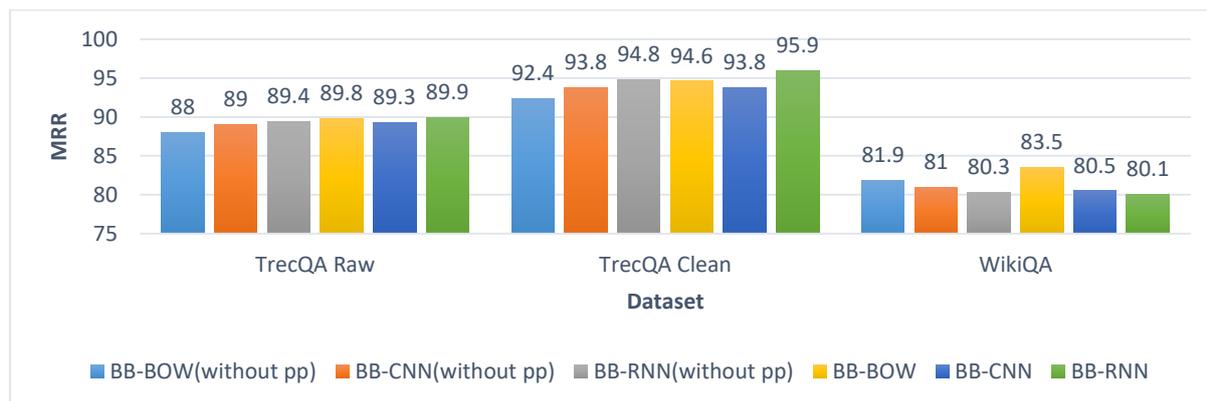

*Fig 17: Effect of lack of the preprocessing on different datasets in terms of MRR*

These results show that the preprocessing has a positive impact on the performance of the language model-based answer-selection models. Given that its impact in some cases is low and even harmful, it is still useful. As mentioned, this may be due to the complexity of the classifiers. But this question needs to be considered separately and the impact of complex classifiers is examined.

## 5.3 The Classifiers Impact

The BERT model has been fine-tuned for various tasks. One of these tasks is sentence classification. BERT model uses the [CLS] token output for the classification task. The model transmits the [CLS] token output vector as the input vector to a fully connected neural network whose output layer presents the classes. This model does not use the output vectors of the other tokens. In this section, the results of BB-BOW, BB-CNN and BB-RNN models are compared with the BB-Baseline model.

*Table 8: Evaluation of the impact of the classifiers on the BAS model*

| Architecture | TrecQA Raw | | TrecQA Clean | | WikiQA | |
|---|---|---|---|---|---|---|
| | MAP | MRR | MAP | MRR | MAP | MRR |
| **BB-Baseline** | 0.869 | 0.886 | 0.908 | 0.942 | 0.789 | 0.810 |
| **BB-BOW** | 0.871 | 0.898 | 0.909 | 0.946 | **0.817** | **0.835** |
| **BB-CNN** | 0.863 | 0.893 | 0.909 | 0.938 | 0.790 | 0.805 |
| **BB-RNN** | **0.872** | **0.899** | **0.915** | **0.959** | 0.784 | 0.801 |

The results are shown in Table 8. These results show that all the classifiers employed instead of the typical Bert classifier improve the performance of the answer-selection model. Using output vectors of all tokens instead of [CLS] token causes the BB-BOW, BB-CNN and BB-RNN models have better performance than the BB-Baseline model. BB-BOW uses the output vector of the other tokens as well as the output vector of the [CLS] token in order to capture more information about the input sentences. In addition to using the output of all tokens, BB-CNN uses the convolutional neural network to overcome the words order problem. In addition to preserving the order of words, BB-RNN also uses recurrent neural network memory to store sentences' information. These results are shown in Figure 18 and Figure 19.

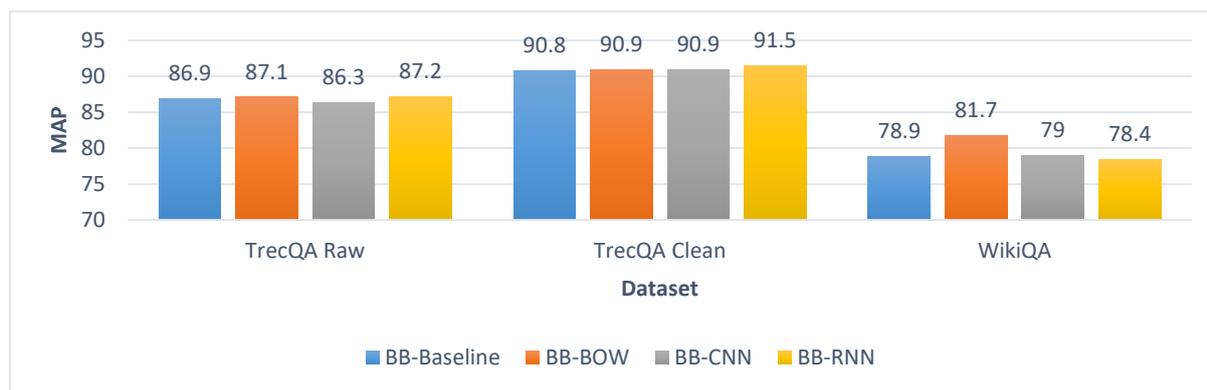

*Fig 18: Model performance with different classifiers on different datasets in terms of MAP*

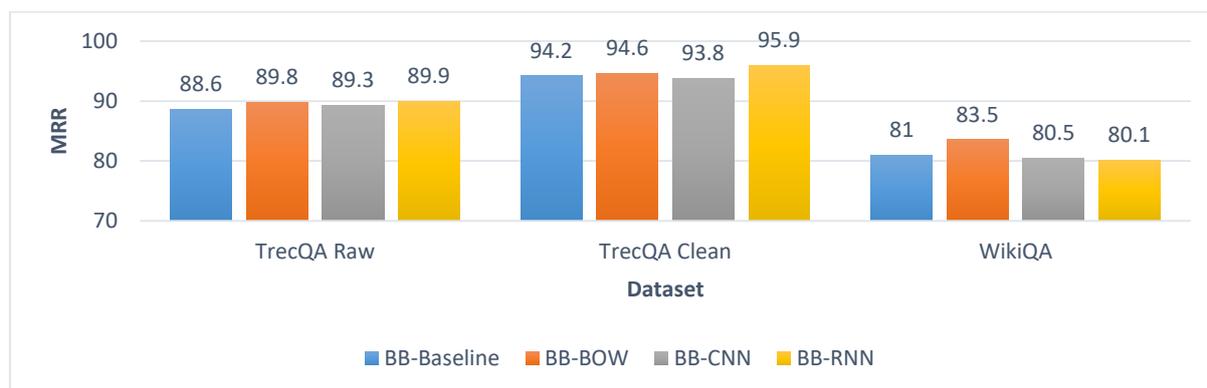

*Fig 19: Model performance with different classifiers on different datasets in terms of MRR*

These results show that using different classifiers instead of the typical classifier of language models can improve the performance of answer-selection models.

## 6. Conclusion

In this research, we present the BAS model, which stands for BERT Answer Selection. This model aims to extract the answer of the user's question from the candidate answers pool and provide it as a final answer to the user. The model consists of three different sections. The first section replaces the EAT tokens with a special token. The second section receives the modified question and answer and generates a representation for each one using the BERT language model. Finally, in the third section, using different classifiers, the relevance matching is calculated. To evaluate our model, we performed several experiments. The experiments were performed on TrecQA Raw, TrecQA Clean and WikiQA datasets.

In the first experiment, the model performance was evaluated. In this experiment, the BAS model was compared with the two baseline models, and it was shown that for the TrecQA Raw and TrecQA Clean datasets, the BB-ROW model was state-of-the-art. BB-BOW model was also state-of-the-art for the WikiQA dataset. The results of the experiment showed that the language model comprehends the sentences better than ordinary neural networks and produces robust representations.

In the second experiment, the effect of the preprocessing section was evaluated. In this experiment, the preprocessing was removed and shown to have a more significant effect on the BB-BOW model than the BB-CNN and BB-RNN models. The results of this experiment showed that the lack of the preprocessing leads to a reduction in the model performance.

In the third experiment, the impact of different classifiers such as BOW, CNN and RNN was evaluated. In this experiment, the performance of the model with the typical BERT classifier was compared with mentioned classifiers in the paper, and it was shown that the BOW, CNN and RNN classifiers performed better than the typical BERT classifiers. The results of this experiment showed that using the output of all tokens instead of [CLS] token lead to a better comprehension of the input sentences.

As a conclusion, we have shown that using strong language models eliminates the need to use knowledge bases and external resources. In other words, if a robust language model such as BERT is employed, the need for additional parts will be eliminated. The reason is the excellent comprehension of language models from languages which makes it easier for the model to identify the relevant answers. The results prove the idea of using the language models. This idea can also be applied to other natural language processing tasks.

As future work, we would like to employ the language models derived from BERT language model. The RoBERTa (Y. Liu et al., 2019) and ALBERT (Lan et al., 2019) are such models. The RoBERTa model has been trained on more data and has provided a more efficient model than the BERT model, which has a better comprehension of the language. Using this model, we can produce more robust representations of input sentences. Instead, the ALBERT model offers a lite BERT model. The reduction of the training parameters in this model has resulted in a reduction in the performance of the BERT model. Instead, it allows us to employ more complex models alongside the language model. Also, we can use more powerful classifiers rather than the classifiers presented in this research.

*Competing interests statement:*

This research did not receive any grant from funding agencies in the public, commercial, or not-for-profit sectors. The authors declare that they have no conflict of interest.